\documentclass[conference]{IEEEtran}
\IEEEoverridecommandlockouts

\usepackage{cite}
\usepackage{amsmath}
\usepackage{amssymb}
\usepackage{amsfonts}
\usepackage{algorithmic}
\usepackage{graphicx}
\usepackage{textcomp}
\usepackage{xcolor}

\def\BibTeX{{\rm B\kern-.05em{\sc i\kern-.025em b}\kern-.08em
    T\kern-.1667em\lower.7ex\hbox{E}\kern-.125emX}}

\usepackage{booktabs} % For formal tables
\usepackage{url}
\usepackage{enumitem}

\newtheorem{definition}{Definition}

\definecolor{gold}{RGB}{255,215,0}
\definecolor{green2}{RGB}{0,128,0}

\begin{document}
% \title{Identify Merchant Category Using\\Credit Card Transactions}
% \title{Merchant Identity Recognition Using\\ Credit Card Transactions}
% \title{Merchant Identity Recognition Using\\ Credit Card Transactions}
\title{Merchant Category Identification Using\\ Credit Card Transactions}
\author{
\IEEEauthorblockN{Chin-Chia Michael Yeh, Zhongfang Zhuang, Yan Zheng, \\ Liang Wang, Junpeng Wang, and Wei Zhang}
\IEEEauthorblockA{\textit{Visa Research} \\
% California, USA \\
\{miyeh, zzhuang, yazheng, liawang, junpenwa, wzhan\}@visa.com}
}

\maketitle

\begin{abstract} % File 1/7
Digital payment volume has proliferated in recent years with the rapid growth of small businesses and online shops.
When processing these digital transactions, recognizing each merchant's real identity (i.e., business type) is vital to ensure the integrity of payment processing systems.
Conventionally, this problem is formulated as a time series classification problem solely using the merchant transaction history.
However, with the large scale of the data, and changing behaviors of merchants and consumers over time, it is extremely challenging to achieve satisfying performance from off-the-shelf classification methods.
In this work, we approach this problem from a \textit{multi-modal learning} perspective, where we use not only the merchant time series data but also the information of \textit{merchant-merchant relationship} (i.e., \textit{affinity}) to verify the self-reported business type (i.e., merchant category) of a given merchant. % Verify vs Double-check
Specifically, we design two individual encoders, where one is responsible for encoding temporal information and the other is responsible for affinity information, and a mechanism to fuse the outputs of the two encoders to accomplish the identification task.
Our experiments on real-world credit card transaction data between 71,668 merchants and 433,772,755 customers have demonstrated the effectiveness and efficiency of the proposed model.
\end{abstract}

\section{Introduction} % File 2/7
From the use of credit cards in local coffee shops to mobile wallet payments in large shopping malls, digital payment systems play an essential role in people's daily routine.
With the growing popularity of cashless payments, new challenges such as detecting identity fraud and falsified merchant identities have surged in recent years.
A system detecting such malicious behavior is now more critical and necessary than ever to protect the integrity of the vast amount of transactions in the payment systems.

\begin{figure*}[t]
    \centering
    \includegraphics[width=0.99\linewidth]{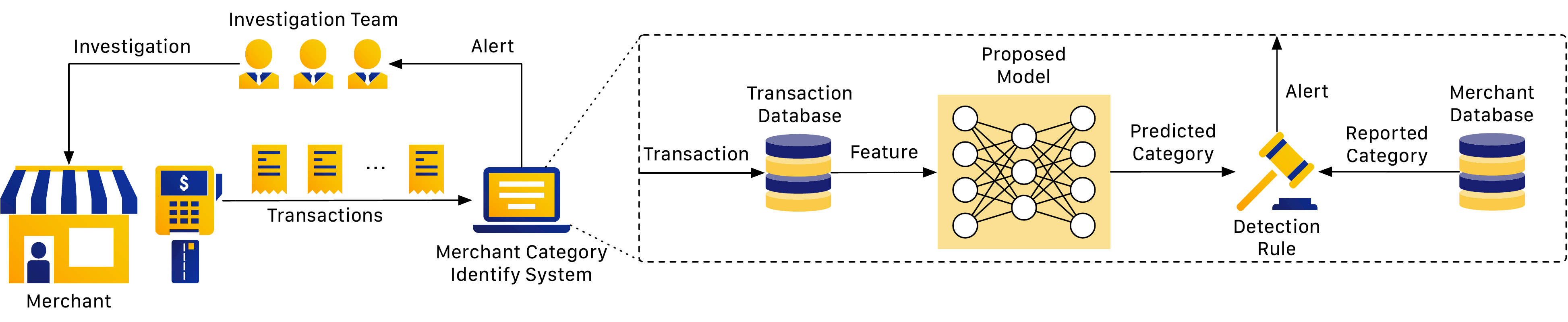}
    \caption{The proposed merchant category identification system.}
    \label{fig:motivation}
\end{figure*}

One specific challenge is to detect \textit{if a merchant falsified its identity} by registering in an incorrect merchant category with payment processing companies.
Accurately identifying every merchant's category has several important implications, where the two most related notable implications are merchant risk level and business legality: \\
\noindent \textbf{Motivating Example 1: Risk Level.} Within payment processing systems, a merchant's risk level is often associated with the merchant's self-reported merchant category.
As a consequence, a high-risk merchant can pretend to be in a low-risk merchant category through reporting fake merchant category to the payment processing company to avoid higher processing fees associated with risky categories. \\
\noindent \textbf{Motivating Example 2: Business Legality.} \textit{Online gambling} is only permitted in some region and territories globally.
Gambling businesses operating covertly in regions and territories where online gambling is prohibited may attempt to register with payment processing companies using business types not relating to gambling to avoid scrutiny from banks and regulators.

Conventionally, there exists a variety of ways to verify a merchant's true identity, such as crawling the merchant's websites, making purchases from the merchant, or physically visiting the merchant.
However, these methods are not scalable in real-world business because of privacy concerns and prohibitively high costs.

In this work, we approach the problem of identifying a merchant's real identity with the help of transaction records.
% Patterns within transaction records of each merchant category are often unique.
As each merchant has a distinct and unique transaction record, a merchant's  \textit{bona fide} identity can be extracted.
Here are examples of \textit{transaction patterns in various time periods}: \\
\noindent \textbf{Motivating Example 3: Transaction Patterns.}
\begin{enumerate}
    \item (Daily) Restaurants' transaction amount typically has two high peaks around lunch and dinner time.
    \item (Weekly) Grocery and department stores tend to be very busy during weekends.
    \item (Yearly) Travel-related merchants (hotels, airlines) are at their peaks during the holiday season.
\end{enumerate}

Merchant category identification problem can be treated as a time series classification problem as merchant transaction histories are represented as time series sequences.
However, off-the-shelf classification methods struggle at producing quality results since not only each merchant category has unique pattern but also each merchant has unique transaction records.
% in a reasonable time frame with the large-scale transaction data incoming at every second.
Additionally, these methods treat each merchant independently without considering the correlations among the merchants.
For example, movie theatre transactions may co-occur with purchases happening in nearby restaurants or ice cream shops in the same shopping mall or city.
These correlations (i.e., \textit{affinity}) help determine a merchant's real identity.

To solve this problem, we propose a novel multi-modal machine learning model that utilizes both the temporal and affinity information from merchants' transaction history data, to improve the scalability and effectiveness of the merchant identification system.
We offer the following contributions:
\begin{itemize}
    \item We formulate the problem of merchant category identification from our real-world business challenges.
    Figure~\ref{fig:motivation} shows how the proposed model fits into the payment industry ecosystem.
    \item We specifically design two sub-networks, namely the temporal encoder and affinity encoder, to respectively handle the temporal and affinity information.
    The outputs of these two sub-networks are fused with a novel mechanism to identify the real category of merchants.
    \item Our experimental studies on real-world merchant transaction data present results that not only demonstrate the effectiveness and efficiency of our model but also showcase how the proposed architecture design effectively captures fraudulent merchants, without introducing many false positives compared with the best performing baseline.
\end{itemize}

\section{Problem Definition} % File 3/7
We begin by defining the features of merchants from the transaction data point of view.
A \textit{merchant} can be represented with two alternative views: \textit{merchant time series} and \textit{merchant affinity vector}.

\begin{definition}[Merchant Time Series]
    {\rm
    A \textit{merchant time series} is a multivariate time series, denoted as $M^{(\mathbf{T})} \in \mathbb{R}^{n \times d}$ where $n$ is the length of the time series and $d$ is the number of dimensions.
    }
    \label{def-mts}
\end{definition}

We extract every merchant's time series from transaction records by performing various feature extraction functions with a non-overlapping sliding window.
Specifically, we use a window of 24 hours (i.e., one day) to extract features shown in Table~\ref{tab:feature} for every day.
We extract $911$ days (i.e., from January 1, 2017, to June 30, 2019) and $10$ features from our merchant transaction dataset. As a result, the time series $M^{(\mathbf{T})}_i$ for any merchant $M_i$ is a $(911, 10)$ sized matrix.

\begin{table}[ht]
\centering
\caption{List of features extracted from the sliding windows of merchant time series.}
\label{tab:feature}
\resizebox{0.99\columnwidth}{!}{
\begin{tabular}{l|l}
\toprule
Feature      & Description \\ \hline
\texttt{numApprTrans} & Number of approved transactions. \\
\texttt{numDeclTrans} & Number of declined transactions. \\
\texttt{numApprCards} & Number of unique cards with approved transactions. \\
\texttt{numDeclCards} & Number of unique cards with declined transactions. \\
\texttt{amtApprTrans} & Volume (amount) of approved transactions. \\
\texttt{amtDeclTrans} & Volume (amount) of declined transactions. \\
\texttt{rateTxnAppr}  & Approval rate. \\
\texttt{rateTxnDecl}  & Decline rate. \\
\texttt{avgAmtAppr}   & Average dollar amount of approved transactions. \\
\texttt{avgAmtDecl}   & Average dollar amount of declined transactions. \\
\bottomrule
\end{tabular}
}
\end{table}

Next, we define \textit{merchant affinity vector}, which is a vector that consists of the similarity between one merchant and every merchant in a merchant set.
\begin{definition}[Merchant Affinity Vector]
    {\rm
    Given a merchant $M$ and a merchant set $\mathbb{M}$ containing $k$ merchants, the \textit{merchant affinity vector} is defined as the similarity of $M$ and each of the $k$ merchants $M^{(A)} \in \mathbb{R}^k$.
    %, where $k$ is the number of merchants in the merchant set~$\mathbb{M}$
    % and $M^{(A)}[j]$ contains the similarity between $M$ and $j$-th merchant in the merchant set~$\mathbb{M}$.
    }
    \label{def-mav}
\end{definition}

The similarity between two merchants can be defined on various features, such as \textit{time series patterns}, \textit{name similarity} and \textit{aggregated numerical features}.
In this work, we define the similarities among merchants based on their interactions with consumers.
Specifically, we count the number of consumers shared by a pair of merchants and use the count as the similarity between them.
Additionally, we query the statistical information within the same period as the merchant time series.

With merchant representation defined, we now focus on the \textit{merchant category}.
The \textit{merchant category} is a set of mutually exclusive classes indicating the business type for a given merchant.

\begin{definition}[Merchant Category]
    {\rm
    Given a merchant $M_i$, we denote the corresponding merchant category as a \textbf{one-hot vector} $Y_i \in \mathbb{B}^c$ where $\mathbb{B} = \{0, 1\}$ and $c$ is the total number of categories.
    % There is only one element in such vector that can be $1$.
    }
    \label{def-mc}
\end{definition}
For example, if merchant $M_i$ is the $j$-th category, $Y_i[j]$ will be $1$ while other elements of the vector will be $0$.
\begin{definition}[Merchant Category Probability]
    {\rm
    $\hat{Y}_i \in \mathbb{R}^c$ (instead of $\mathbb{B}^c$) denotes the predicted probability for merchant $i$ with respect to each category. $\sum_{j\in c} \hat{Y}_i[j] = 1$.
    }
\end{definition}

Built on top of previous definitions, our merchant category identification problem is defined as:
\begin{definition}
    {\rm
    Given a set of trustworthy merchants~$\mathbb{M}_t$ with their features $R(\mathbb{M}_t)$ and their corresponding set of merchant categories~$\mathbb{Y}_t$, the goal of the  \textit{merchant category identification problem} is to learn a model~$\mathsf{F}$ from $(\mathbb{M}_t, \mathbb{Y}_t)$, which can be used to identify the \textit{true} merchant category of a suspicious merchant.
    Model~$\mathsf{F}$ is learned by minimizing the following loss function:
    \begin{equation}
    \sum_{M_i \in \mathbb{M}_t, Y_i \in \mathbb{Y}_t} \texttt{loss}(\mathsf{F}(R(M_i)), Y_i)
    \end{equation}
    where the $\texttt{loss}(\cdot)$ is the negative log likelihood loss function because merchant categories are mutually exclusive. In this study, we use $R(M_i) = \big\{ M_i^{(\mathbf{T})}, M_i^{(A)} \big\}$.
    }
    \label{def-mci}
\end{definition}

\section{Model Architecture} % File 4/7
\begin{figure}[ht]
    \centering
    \includegraphics[width=0.75\columnwidth]{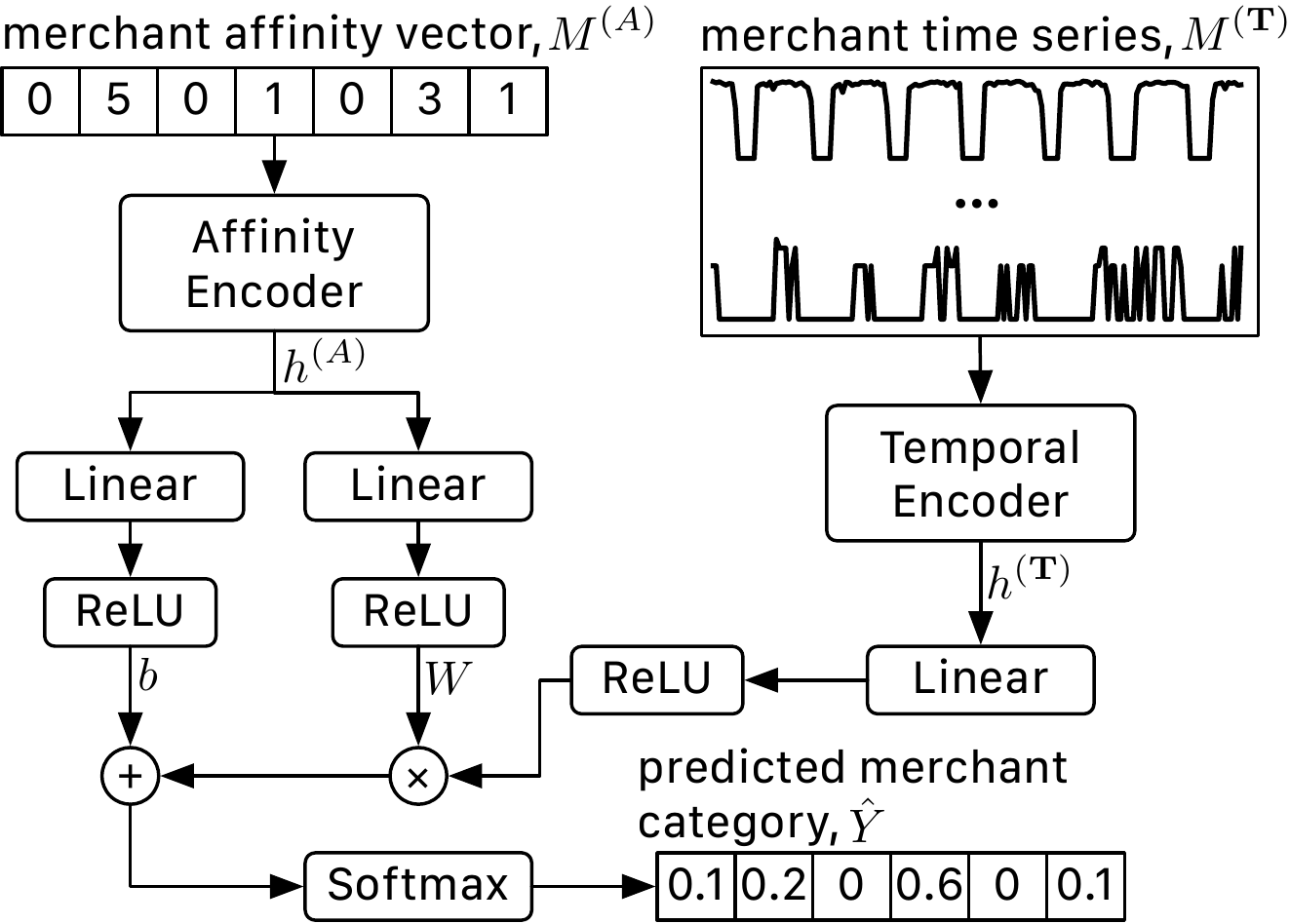}
    \caption{Overall architecture design. The detailed design of Affinity Encoder and Temporal Encoder is shown in Figure~\ref{fig:affenc} and Figure~\ref{fig:temenc}, respectively.}
    \label{fig:overall}
\end{figure}

In this section, we describe the proposed model architecture.
We use Figure~\ref{fig:overall} to depict an example of the forward pass of our proposed model.
The inputs are the merchant affinity vector~$M^{(A)} \!\in\! \mathbb{R}^k$ and merchant time series~$M^{(\mathbf{T})} \!\in\! \mathbb{R}^{n \times d}$ of a given merchant and the output is the predicted merchant category probability~$\hat{Y} \!\in\! \mathbb{R}^c$ for the merchant, where $n$ is the length of the time series, $d$ is the dimensionality of the time series, $k$ is the number of known merchants, and $c$ is the number of merchant categories.

The framework has two independent encoders for learning the hidden representations of the merchant time series and merchant affinity vector.
The merchant affinity representation~$h^{(A)} \in \mathbb{R}^{n_k}$ and merchant temporal representation~$h^{(\mathbf{T})} \in \mathbb{R}^{n_k}$ are extracted using $M^{(A)}$ and $M^{(\mathbf{T})}$ in parallel with the corresponding encoders, where $n_k$ is the dimension of the hidden representations.
\begin{equation*}
\begin{split}
    h^{(A)} & \gets \texttt{AffinityEncoder}(M^{(A)}) \\
    h^{(\mathbf{T})} & \gets \texttt{TemporalEncoder}(M^{(\mathbf{T})}) \\
\end{split}
\end{equation*}
The size of $h^{(A)}$ and $h^{(\mathbf{T})}$ depends on the hyper-parameter settings of  $\texttt{AffinityEncoder}$ and $\texttt{TemporalEncoder}$.
The details about each encoders are outlined in Section~\ref{sec:affenc} and Section~\ref{sec:temenc}.
In our particular implementation, both representations are sized $64$ vectors (i.e., $n_k = 64$).
The merchant temporal representation~$h^{(\mathbf{T})}$ is further processed by a $\texttt{Linear}$ layer and a $\texttt{ReLU}$ layer.
We denote this intermediate temporal representation as~$\bar{h}^{(\mathbf{T})} \in \mathbb{R}^{n_k}$.
\begin{equation*}
    \bar{h}^{(\mathbf{T})} \gets \texttt{ReLU}(\texttt{Linear}(h^{(\mathbf{T})}))
\end{equation*}
The $\texttt{Linear}$ layer used to process~$h^{(\mathbf{T})}$ has both its input and output size set to $n_k$.

In parallel, the merchant affinity representation~$h^{(A)}$ is used as the input to two independent linear layers to generate the weight matrix~$W \in \mathbb{R}^{n_k \times c}$ and bias vector~$b \in \mathbb{R}^c$.
\begin{equation*}
\begin{split}
    W & \gets \texttt{ReLU}(\texttt{Linear}(h^{(A)})) \\
    b & \gets \texttt{ReLU}(\texttt{Linear}(h^{(A)})) \\
\end{split}
\end{equation*}
The input and output size for the \texttt{Linear} layer responsible for generating $W$ are $n_k$ and $n_k \times c$; the input and output size for the \texttt{Linear} layer responsible for generating $b$ are $n_k$ and $c$.
%In most deep learning package, the linear layer typically generated vector instead of a matrix, we reshape the output $n_k c$ sized vector to the matrix with shape $(n_k, c)$.
Lastly, the predicted merchant category probability~$\hat{Y}$ is computed as follows:
\begin{equation*}
    \hat{Y} \gets \texttt{Softmax}(\bar{h}^{(\mathbf{T})} W + b)
\end{equation*}

We build the model architecture design upon the assumption that merchants targeting different ``types'' of consumers could have very different behavior in terms of temporal representation even if these merchants belongs to the same category.
For example, grocery stores targeting different market segments exhibit distinct transaction volume dynamics due to the discrepancy in consumers' behaviors from different market segments (e.g., urban versus suburban).
By considering the $W$ and $b$ together with the $\texttt{Softmax}$ layer, we essentially generate a logistic regression model from the merchant affinity representation~$h^{(A)}$.
As the merchant affinity representation captures the difference between merchants in terms of consumers' tastes, the proposed model generates different logistic regression models for different ``types'' of merchants.

\subsection{Affinity Encoder}
\label{sec:affenc}

\begin{figure}[ht]
    \centering
    \includegraphics[width=0.75\linewidth]{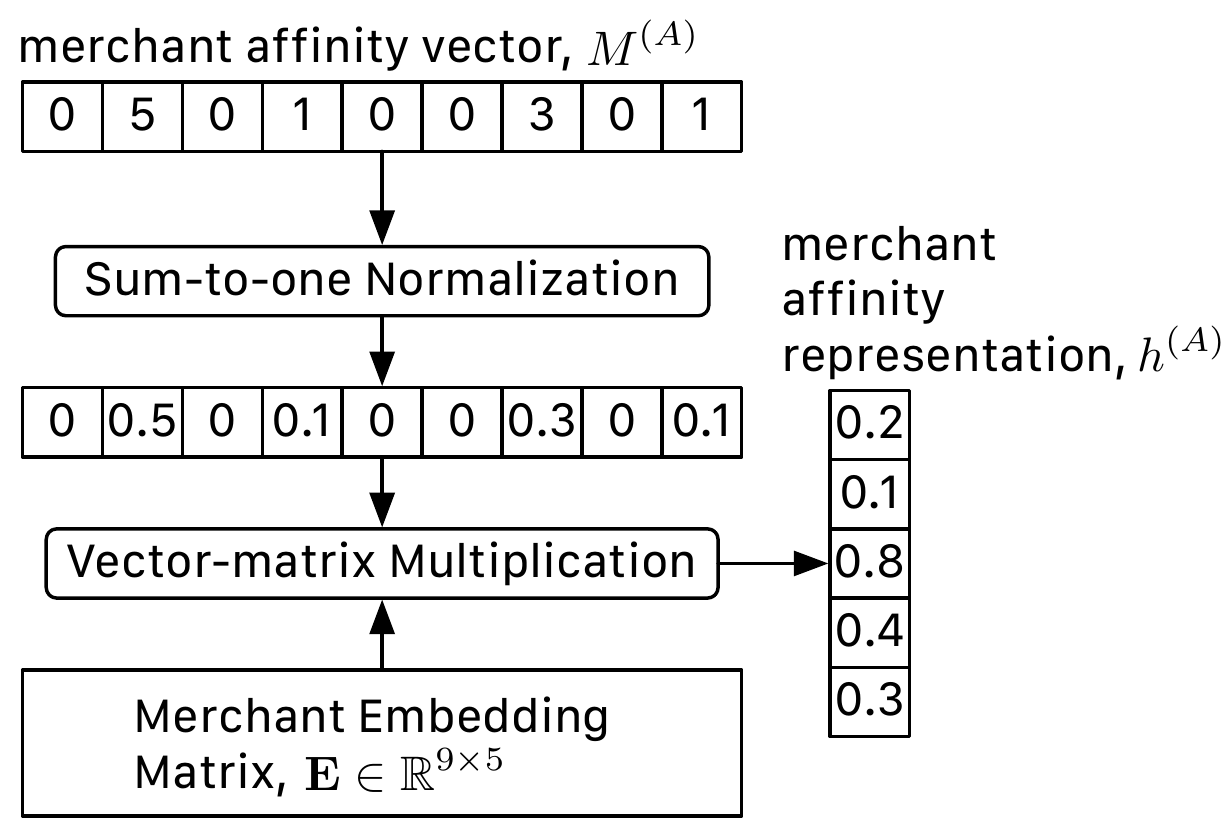}
    \caption{Architecture design for the affinity encoder. In this example, there are $9$ known merchants and the size of embedding is set to $5$.}
    \label{fig:affenc}
\end{figure}

The affinity encoder uses a simple embedding-based model design, as shown in Figure~\ref{fig:affenc}.
The input to the model is the merchant affinity vector~$M^{(A)}$, which describes a given merchant's relationship with a set of known merchants (i.e., the number of shared customers).
Note, the merchant described by~$M^{(A)}$ may or may not be in the set of known merchants.
There is only one trainable variable $\mathbf{E}$, which consists of embeddings for the set of known merchants.

We first normalize the input affinity vector~$M^{(A)}$ with the following equation:
\begin{equation*}
    \bar{M}^{(A)} \gets \frac{M^{(A)}}{\sum_i M^{(A)}[i]}
\end{equation*}
The main reason for applying this normalization step is to make sure the output of the merchant affinity encoder does not produce representations with extreme values while the input merchant has a much higher transaction volume (i.e., much higher $L1$ vector norm) when compared to other merchants.
In our dataset, the $L1$ norm for merchant affinity vectors ranges from $64$ to $701,464,302$.

The output of the merchant affinity is computed using:
\begin{equation*}
    h^{(A)} \gets \bar{M}^{(A)}\textbf{E}
\end{equation*}
Essentially, the merchant affinity representation~$h^{(A)}$ is a weighted sum of the merchant embeddings stored in $\textbf{E}$.
As the affinity is defined by the number of shared consumers between merchants, $h^{(A)}$ of a given merchant depends on what other merchants were visited by the previously mentioned merchant's consumers.
In other words, the $h^{(A)}$ of a given merchant captures the ``taste" of the merchant's consumers.
Since a merchant ``does not depend'' on itself, the affinity between a merchant and itself is $0$. Therefore, the affinity encoder is also applicable to merchants that are not available during the training time.
%The affinity between a merchant and itself is defined as $0$, which means that the $h^{(A)}$ of a merchant does not depend on itself; therefore, the affinity encoder can be applied to a merchant that is not available during training time.
The affinity encoder design can be considered as a special case of inductive representation learning on graphs~\cite{hamilton2017inductive}: the set of merchant affinity vector can be treated as the adjacency matrix of a bipartite graph~\cite{hamilton2017inductive}, which generates embeddings by aggregating features from a node’s local neighborhood.

In our implementation, as we set the embedding size to $64$, the size of the embedding matrix~$\mathbf{E}$ is $(43000, 64)$, when the number of merchants in the training data is $43,000$.
The vector-matrix multiplication is implemented in a sparse fashion for its improved efficiency as most of the merchant affinity vectors found in our datasets are sparse.
To further reduce the memory usage of the training process, we limit the number of non-zero values to be at most $8,192$ in the vector-matrix multiplication.
% In the case where there are more than $8,192$ non-zero values, we use the larger $8,192$ non-zero values.

\subsection{Temporal Encoder}
\label{sec:temenc}

\begin{figure}[t]
    \centering
    \includegraphics[width=0.90\columnwidth]{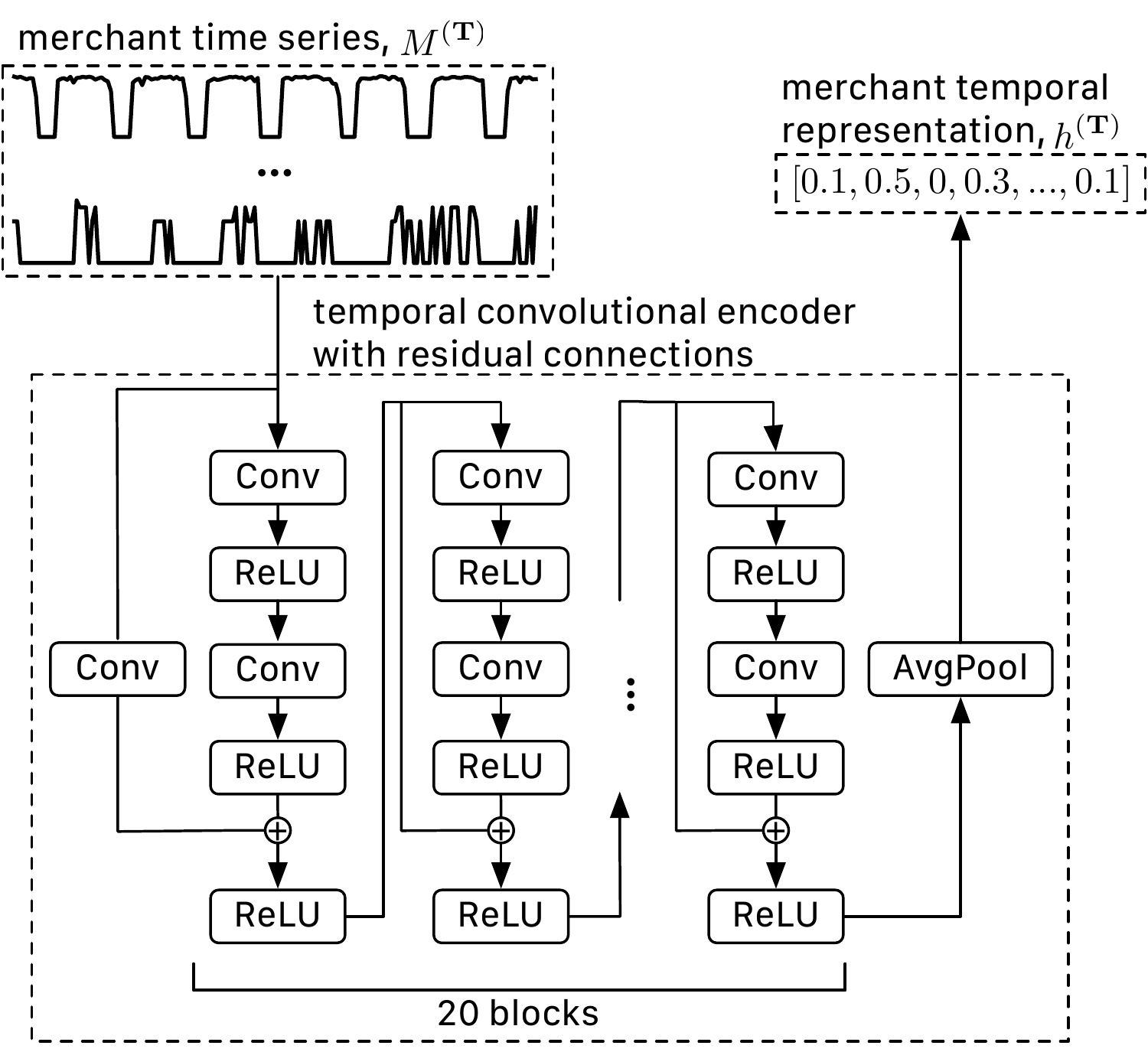}
    \caption{Architecture design of temporal encoder. All the $\texttt{Conv}$ layers are $1D$ convolutional layers, and the $\texttt{AvgPool}$ layer is $1D$ average pooling layer.}
    \label{fig:temenc}
\end{figure}

The design of the temporal encoder (see Figure~\ref{fig:temenc}) is inspired by the Temporal Convolutional Network (TCN)~\cite{bai2018empirical}.
Specifically, the temporal encoder uses $20$ stacks of the residual block presented in~\cite{bai2018empirical} to extract the merchant temporal representation $h^{(\mathbf{T})}$ of a given merchant time series $M^{(\mathbf{T})}$.

For each residual block, the main passage consists of four layers (i.e., $\texttt{Conv-ReLU-Conv-ReLU}$) and the residual connection consists of one optional layer (i.e., $\texttt{Conv}$).
The residual connection merges with the main passage by element-wise addition, then passing trough another $\texttt{ReLU}$ before passing to subsequent layers.
The number of channels for all $\texttt{Conv}$ layers is set to $64$.
The receptive field size is set to $3$ for $\texttt{Conv}$ layers in the main passage and $1$ for the optional $\texttt{Conv}$ layer.
The optional $\texttt{Conv}$ layer is applied when the input and output dimensionality of the residual block do not match; therefore, it is only applied in the first residual block (i.e., $d = 10$ in our system).

To produce the merchant temporal representation vector $h^{(\mathbf{T})}$, a global average pooling (i.e., $\texttt{AvgPool}$) layer is applied to the output of the last residual block, which summarized the hidden representation across time.
Under our specific problem and hyper-parameter settings, the input to the $\texttt{AvgPool}$ layer for each merchant is a sized $(911, 64)$ matrix and the resulting output is a sized $64$ vector.
Dropout layers are used but omitted to simplify Figure~\ref{fig:temenc}.
The omitted dropout layers are positioned after the two $\texttt{ReLU}$ in the main passage of each residual block.
Unlike the original TCN~\cite{bai2018empirical}, we do not use weight normalization.
Because our goal is classification instead of sequence modeling, we use a regular convolutional layer instead of a causal convolutional layer.

\section{Experiments} % File 5/7
\noindent \textbf{Datasets.} The dataset consists of transactions involving 71,668 merchants and 433,772,755 customers from January 1, 2017, to June 30, 2019.
We collect 56 merchant categories for this work, and the numbers of merchants in different categories are mostly imbalanced:
the most popular category (i.e., the \textit{fast food} category has 7,555 merchants) is 33 times larger than the least popular category (i.e., the \textit{hospital} category with 226 merchants) in this dataset.
We use stratified 5-fold cross-validation in our experiment.
The dataset is divided into five folds by merchants.
Each of the first three folds out of the five folds consists of $14,334$ merchants, while each of the last two folds consists of $14,333$ merchants.
For each repetition, we use three folds as the training set, one fold as the validation set, and one fold as the test set.
% We use the training set exclusively for training the model.
% The validation set used for hyperparameter tuning and model selection.
% We evaluate the performance of the model using the test data.

\noindent \textbf{Metrics.} To measure the performance of different methods, we use the following performance metrics: micro f1 score (Micro F1), macro f1 score (Macro F1), average rank (AR), top 3 hit rate (Hit@3), and top 5 hit rate (Hit@5).
%The average rank is computed by first ordering the merchant categories based on estimated probability for each merchant, then we have the rank of each merchant's true category, the AR is finally computed by averaging across all the merchants' rank of true category.
We compute the average rank using the following procedures: we first order the merchant categories based on the estimated probabilities for each merchant; then, the AR is computed by averaging across all the merchants' rank of true category.
We compute the averaged value for each performance metric over the $5$ repetitions in the cross-validation.

\noindent \textbf{Pre-processing.} For each merchant, we extract the daily $10$ features shown in Table~\ref{tab:feature} to form the time series representation~$M^{(\mathbf{T})}_i \in \mathbb{R}^{911 \times 10}$.
To extract the merchant affinity vector, for every merchant, we compute the number of shared customers \textit{only} with each merchant in the training set; therefore, none of the merchants in the test or validation set has a merchant embedding in $\textbf{E}$.
Because the number of merchants in each fold is not the same, the size of merchant affinity vector $M^{(A)}_i$ could be $43,002$, $43,001$, or $43,000$ depending on the number of merchants in the training set.
To process the amount of data in time, we extract both merchant time series representations and merchant affinity vectors using Hive~\cite{hive}.

\noindent \textbf{Implementation.} In terms of optimization, we use Stochastic Gradient Descent with one cycle learning rate policy~\cite{smith2018disciplined}~\cite{smith2019super}, and we use the validation loss to select the hyper-parameters, i.e., dropout, learning rate, weight decay, and momentum.
We train each model for 128 epochs, and at the end of each epoch, we saved a checkpoint model.
We select the best model with the lowest validation cost from available checkpoint models for testing.
We implement all models, \textit{except} the nearest neighbor models, with PyTorch~\cite{paszke2019pytorch}.
For the nearest neighbor models, we use the implementation from tslearn~\cite{tslearn}.
We conduct our experiments on a Red Hat Linux server with Intel Xeon E5-2650 v4 CPU and Nvidia Tesla P100 GPU.

\subsection{Temporal Encoder Alternative Design}
\label{sec:tempenc}
\begin{figure}[ht]
    \centering
    \includegraphics[width=0.70\linewidth]{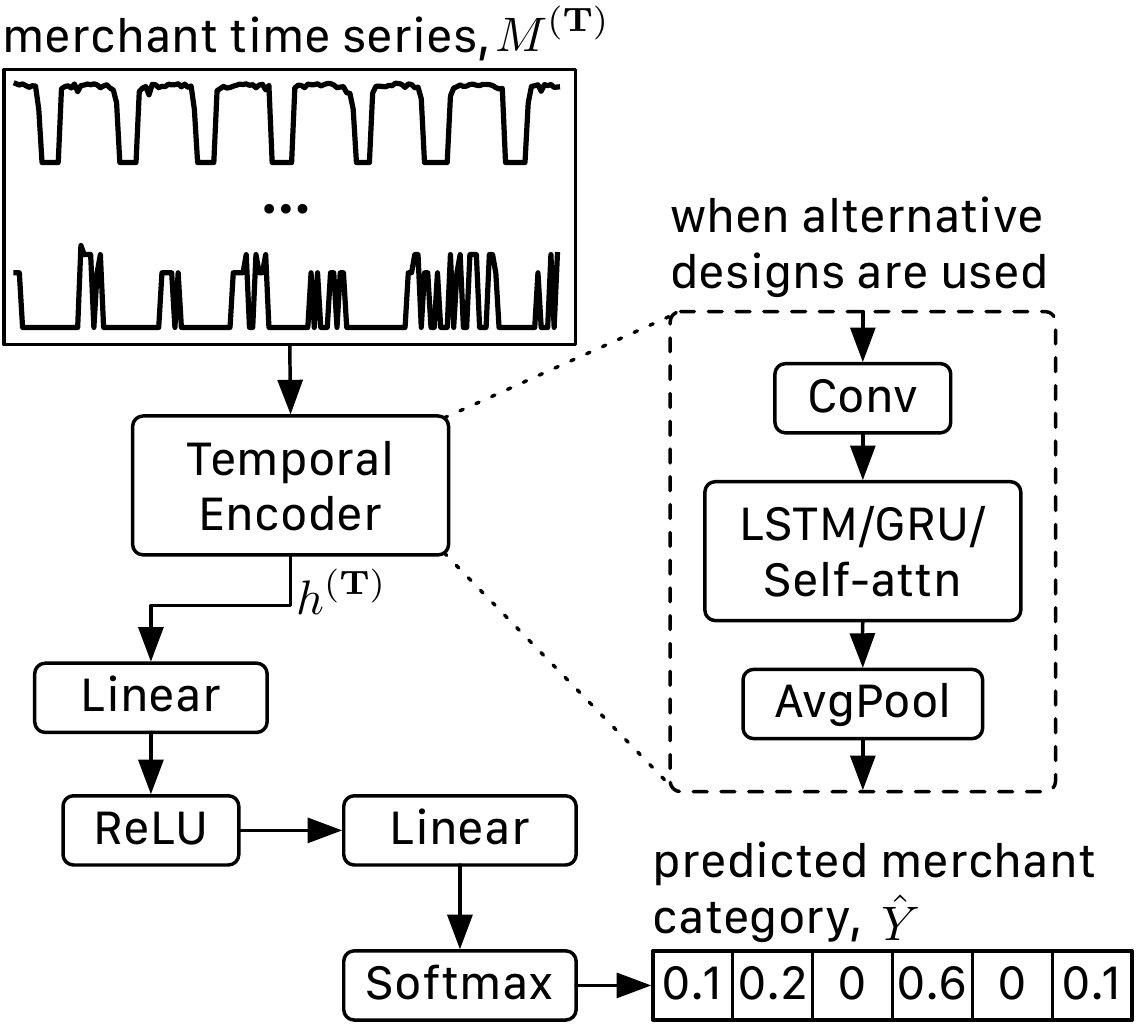}
    \caption{Architecture design used for evaluating different temporal encoders.}
    \label{fig:temexp}
\end{figure}

Convolution-based model is only one of the options for modeling time series data.
In this section, we examine the possibility of using alternative architectures, like Long short-term memory (LSTM)~\cite{hochreiter1997long}, Gated recurrent unit (GRU)~\cite{cho2014properties}, and Self-attention module (Self-attn)~\cite{vaswani2017attention} for temporal encoder.
We use the architecture shown in Figure~\ref{fig:temexp} for the experiment.

In this study, we do not use affinity encoder because we want to focus our study on the temporal encoder.
We first feed the merchant time series~$M^{(\mathbf{T})}$ to the temporal encoder; then, we pass the output temporal representation through $\texttt{Linear-ReLU-Linear-Softmax}$ to get the merchant category estimation.
When the alternative design is used, we first feed $M^{(\mathbf{T})}$ into a $1D$ $\texttt{Conv}$ layer with a receptive field of size 3 and 64 channels to extract local features similar to how $\texttt{Conv}$ layer is applied before LSTM in~\cite{peters2018deep}.
Then, either the LSTM, GRU, or the Self-attn layer are applied to the output of $\texttt{Conv}$ layer.
For both LSTM and GRU, we use the bi-direction variant with $64$ channels and set the number of layers to $2$.
We use the ``multi-head'' attention variant for the self-attention. Specifically, we use $4$ heads, $64$ channels, and $3$ layers in this module.

In addition to the three crucial deep learning-based baselines, we also provide three more basic baselines: random guess (Random), one nearest neighbor with Euclidean distance (1NN), and logistic regression (LR).
The experiment result is summarized in Table~\ref{tab:temp_acc}.
% $\ddagger$ and $\dagger$ indicate that the baseline is significantly worse than the proposed temporal encoder with a significance level of $1\%$ and $5\%$, respectively.
For average rank(AR), the optimal value is $1$, thus the smaller the number, the better the results.
While for other metrics, the larger the number, the better the results.

\begin{table}[ht]
\centering
\caption{Comparing to different temporal encoder designs. $\ddagger$ and $\dagger$ indicate that the baseline is worse than the proposed model significantly (i.e., two sample t-test) with significance level of $1\%$ and $5\%$, respectively.}
\label{tab:temp_acc}
% \resizebox{0.95\columnwidth}{!}{
\begin{tabular}{l|ccccc}
\toprule
          & Micro F1 & Macro F1 & AR      & Hit@3  & Hit@5  \\ \hline
Random    & 0.0181$\ddagger$   & 0.0151$\ddagger$   & 28.4897$\ddagger$ & 0.0539$\ddagger$ & 0.0896$\ddagger$ \\
1NN       & 0.4667$\ddagger$   & 0.3949$\ddagger$   & 15.8113$\ddagger$ & 0.4877$\ddagger$ & 0.4974$\ddagger$ \\
LR        & 0.5483$\ddagger$   & 0.4537$\ddagger$   & 4.0877$\ddagger$  & 0.7388$\ddagger$ & 0.8167$\ddagger$ \\ \hline
GRU       & 0.7640$\ddagger$   & 0.6590$\dagger$   & 2.2289  & 0.8880$\dagger$ & 0.9242 \\
LSTM      & 0.7156$\ddagger$   & 0.5917$\ddagger$   & 2.5277$\ddagger$  & 0.8603$\ddagger$ & 0.9046$\ddagger$ \\
Self-attn & 0.6603$\ddagger$   & 0.5048$\ddagger$   & 2.8631$\ddagger$  & 0.8281$\ddagger$ & 0.8825$\ddagger$ \\ \hline
\textbf{Proposed}  & 0.7714   & 0.6666   & 2.1999  & 0.8921 & 0.9269 \\
\bottomrule
\end{tabular}
% }
\end{table}

The best performing baseline is LR.
81\% of all times, the correct merchant category is within the top 5 most probable categories.
For the deep learn-based baseline methods, GRU is the most accurate method.
GRU outperforms all other baseline methods in all performance metrics.
Comparing with LR, GRU pushes all metrics (except AR) by at least 0.1.
The improvement is not only from the deep learning model but also from treating the data as a time-series sequence, other than a non-ordered high dimensional data matrix.
When comparing the proposed temporal encoder architecture with baselines, we can see that it almost outperforms all baselines significantly on all performance metrics.
The only exception is GRU.
Although the proposed method does outperform GRU on all performance metrics, but only rejects the null hypothesis in the significance test with Micro F1, Macro F1, and Hit@3.

We further compare the runtime difference between different deep learning-based temporal encoder designs.
The average training time per iteration and test time per merchant for each encoder design are shown in Table~\ref{tab:temp_time}.
The proposed method is the most efficient one, uses less than $1/3$ of GRU runtime.

\begin{table}[ht]
\centering
\caption{Runtime of different temporal encoder designs.}
\label{tab:temp_time}
% \resizebox{0.85\columnwidth}{!}{
\begin{tabular}{l|ccc|c}
\toprule
        & GRU    & LSTM   & Self-attn & \textbf{Proposed} \\ \hline
train time (sec) & 0.270 & 0.219 & 0.093    & 0.082  \\
test time (ms) & 1.851 & 1.600 & 0.792    & 0.378  \\
\bottomrule
\end{tabular}
% }
\end{table}

Because both Self-attn and the proposed method (i.e., $\texttt{Conv}$ layer) do not perform back-propagation through time, it is much faster than recurrent-based methods (i.e., GRU and LSTM).
Although GRU might not be significantly worse than the proposed method, because of the large margin between the training runtime, the proposed method is still the preferred method over the alternatives.

The experiment conducted for this section shows that the particular temporal encoder design we proposed outperforms other commonly seen deep learning-based models in both accuracy and speed.

\subsection{Affinity Encoder: Time-Memory and Accuracy Trade-off}
\begin{figure}[ht]
    \centering
    \includegraphics[width=0.70\linewidth]{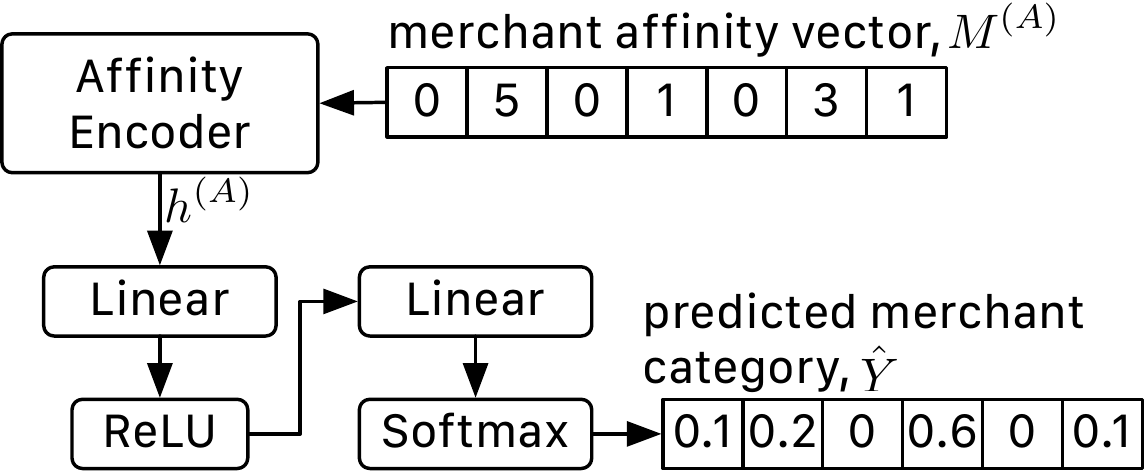}
    \caption{Architecture design used for evaluating just affinity encoders.}
    \label{fig:affexp}
\end{figure}

In this section, we focus on exploring the possibility of reducing the time-memory cost associated with the affinity encoder.
Similar to Section~\ref{sec:tempenc}, we also use a simplified overall architecture that only uses the merchant affinity vector as its input.
The architecture used in this experiment is shown in Figure~\ref{fig:affexp}.

\begin{figure}[ht]
    \centering
    \includegraphics[width=0.90\linewidth]{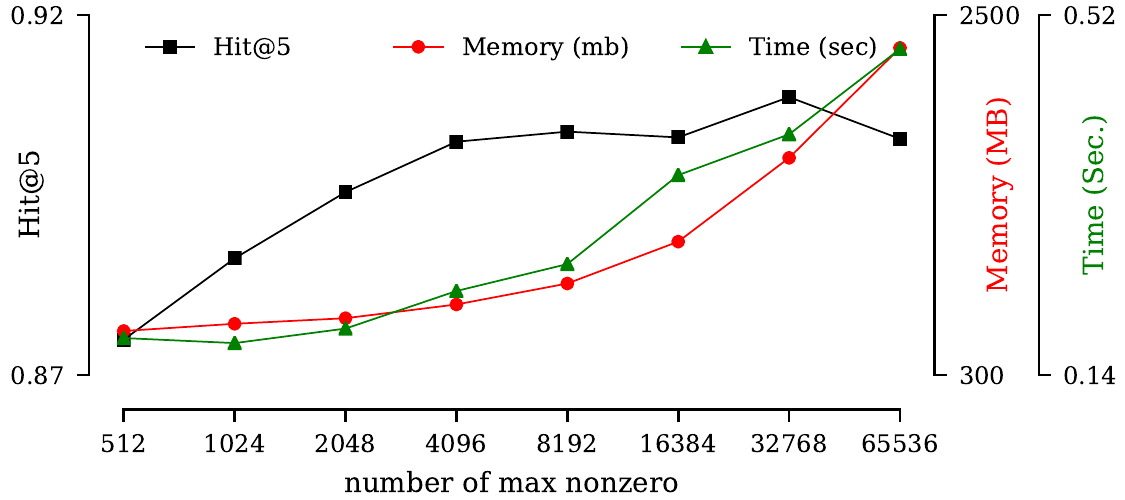}
    \caption{Time-memory and accuracy (Hit@5) trade-off for affinity encoder.}
    \label{fig:afftrade}
\end{figure}

In the affinity encoder, merchant affinity representation~$h^{(A)}$ is extracted from the merchant affinity vector~$M^{(A)}$ ($h^{(A)} \gets \bar{M}^{(A)}\textbf{E}$), and then it is inputted to a two-layer model (i.e., $\texttt{Linear-ReLU-Linear-Softmax}$) to compute the probability for each merchant's category.
To study the trade-off between time-memory cost and accuracy, we focus our study on the effect of varying the sparsity of the merchant affinity vector.
As the merchant affinity vector is stored and used as a sparse vector, increasing its sparsity can dramatically reduce the time-memory cost of both storing and processing it since less non-zero values need to be stored, and fewer parameters in $E$ need to be updated.
Specifically, we increase the sparsity by removing the smaller values from~$M^{(A)}$ and only retaining the largest~$\bar{k}$ values.
If the number of non-zero values in~$M^{(A)}$ for a merchant is smaller than~$\bar{k}$, we keep all the values.
We vary~$\bar{k}$ from $512$ to $65,536$ and measure the averaged runtime per iteration, GPU memory usage, and various performance measurements for accuracy.
For example, Figure~\ref{fig:afftrade} shows how the averaged runtime, memory usage, and Hit@5 change as~$\bar{k}$ is varied.
We put them on the same figure with the same x-axis and different y-axis.
Because the conclusion remains the same with the other accuracy performance measurements, we omit the other metrics in the figure for brevity.
Note, the x-axis is plotted in logarithmic scale.

Both memory and time grow with~$\bar{k}$ linearly as expected, but Hit@5 plateaus when~$\bar{k}$ reaches $4,096$.
As a result, we choose to set~$\bar{k}$ to $8,192$, which is the next tested~$\bar{k}$ after the Hit@5 plateaus, for the full experiment.
Note, although the particular hardware we use is capable of handling the raw merchant affinity vector~$M^{(A)}$ for the particular dataset we are using, such time-memory and accuracy trade-off study is still necessary as the number of merchants in the real dataset is several orders of magnitude larger than the dataset we present in this paper.

\subsection{Affinity Encoder Visualization}
\label{sec:affvis}
\begin{figure}[ht]
    \centering
    \includegraphics[width=0.99\linewidth]{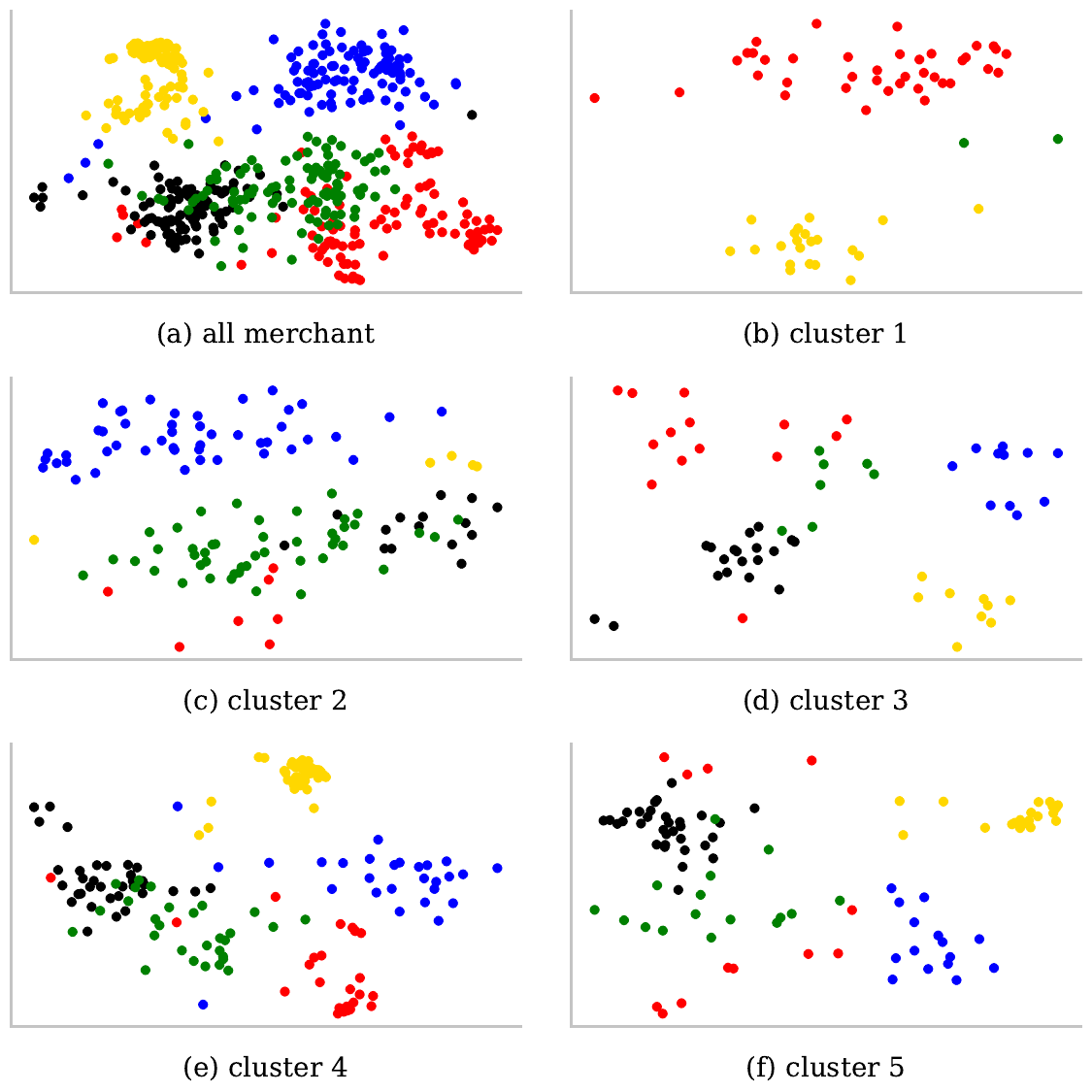}
    \caption{Distribution of merchant temporal representations has better class separation when clustered using merchant affinity representation. The color of the marker indicates the merchant's category (i.e., \textcolor{red}{grocery}, fast food, \textcolor{blue}{barber}, \textcolor{gold}{gas station}, and \textcolor{green2}{restaurant}).}
    \label{fig:emb}
\end{figure}

To demonstrate the effectiveness of the affinity encoder, we visualize the merchant temporal representations (i.e., $h^{(\mathbf{T})}$) of $500$ merchants from the $5$ most populated merchant categories (i.e., $100$ merchants per category) under two scenarios.
The visualization (i.e., projection in $2-D$ space) for each $h^{(\mathbf{T})}$ is generated using multidimensional scaling~\cite{kruskal1964multidimensional} and shown in Figure~\ref{fig:emb}.

In the first scenario, we simply visualize all $500$ merchants' $h^{(\mathbf{T})}$ in the same scatter plot (Figure~\ref{fig:emb}.a).
The separation between different merchant categories is reasonably good, but \textcolor{green2}{restaurants} overlap with other food related categories (i.e., \textcolor{red}{grocery} and fast food).
For the second scenario, we first cluster the merchants using merchant affinity representations (i.e., $h^{(A)}$) as the intuition behind the model is that we want to distinguish merchants with a similar consumer base (i.e., similar $h^{(A)}$).
We arbitrarily choose to cluster the merchants into $5$ clusters with the $k$-means algorithm~\cite{lloyd1982least}.
Next, we visualize merchants belonging to each of the clusters independently, as shown in Figure~\ref{fig:emb}.b to Figure~\ref{fig:emb}.f, and from the figures, the overlapping between different food-related categories is reduced comparing to Figure~\ref{fig:emb}.a.
As a result, it is demanded to fuse the affinity model with the temporal mode.

\subsection{Temporal-affinity Fusion Mechanism}
\begin{figure}[ht]
    \centering
    \includegraphics[width=0.80\linewidth]{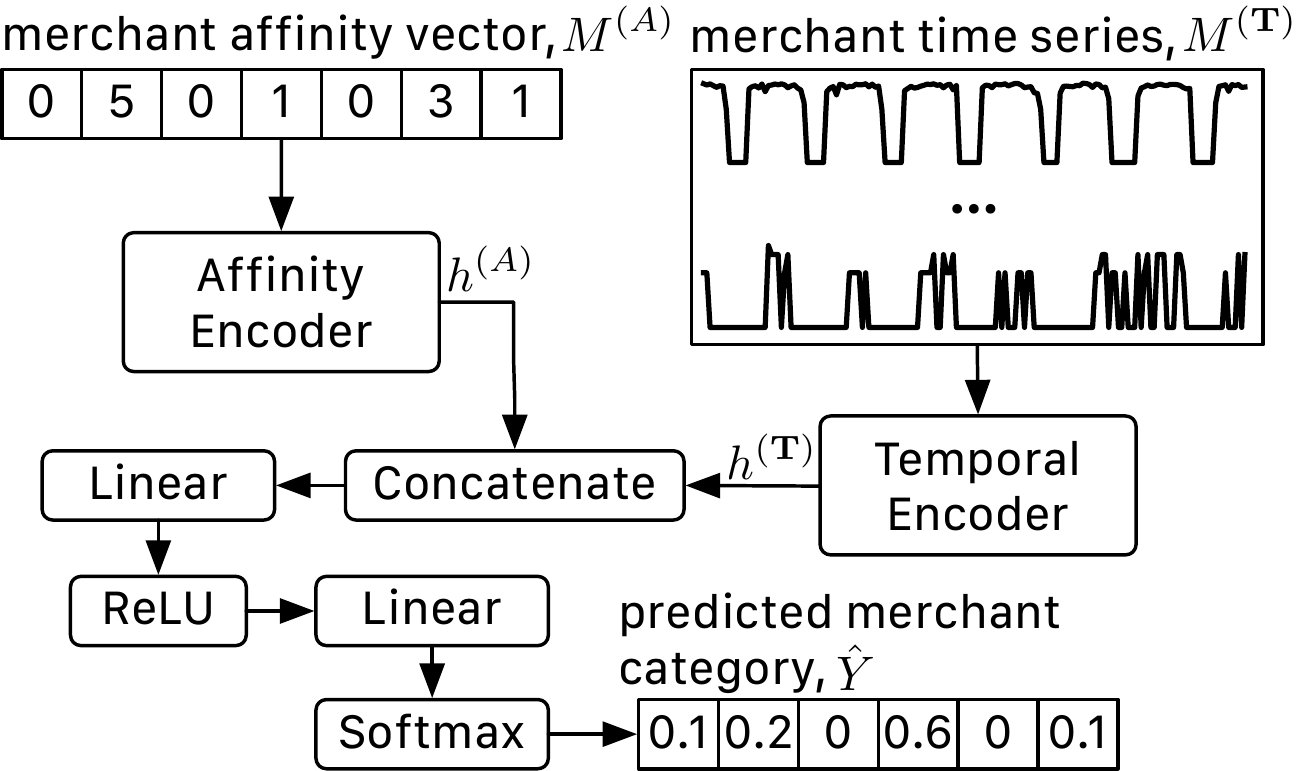}
    \caption{Alternative design for combining merchant affinity representation with merchant time series representation.}
    \label{fig:simple}
\end{figure}

Comparing to the proposed architecture shown in Figure~\ref{fig:overall}, a more straightforward way to fuse both merchant affinity representation and merchant time series representation is to simply concatenate the vectors.
Figure~\ref{fig:simple} shows the simplified architecture.

In this section, we compare the proposed architecture (i.e., Figure~\ref{fig:overall}) with the alternative architecture (i.e., Figure~\ref{fig:simple}) to demonstrate the merit of the proposed architecture.
Table~\ref{tab:fusion} shows the performance of both fusion architectures.
We also put the temporal-encoder-only and affinity-encoder-only models as references to show how fusing both temporal, and affinity representations improve the performance.
The temporal-encoder-only and affinity-encoder-only models are trained using the architectures shown in Figure~\ref{fig:temexp} and Figure~\ref{fig:affexp}, respectively.

\begin{table}[ht]
\centering
\caption{Comparing proposed model to a simple-concatenate design (Simple) along with temporal-encoder-only (Temporal) and affinity-encoder-only (Affinity) models.}
\label{tab:fusion}
% \resizebox{0.85\columnwidth}{!}{
\begin{tabular}{l|ccccc}
\toprule
         & Micro F1 & Macro F1 & AR     & Hit@3  & Hit@5  \\ \hline
Temporal & 0.7714   & 0.6666   & 2.1999 & 0.8921 & 0.9269 \\
Affinity & 0.6511   & 0.5401   & 2.5640 & 0.8463 & 0.9038 \\ \hline
Simple   & 0.7797   & 0.6790   & 2.1119 & 0.8985 & 0.9321 \\
% \textbf{Proposed} & 0.8084   & 0.7254   & 1.8589 & 0.9212 & 0.9481 \\
\textbf{Proposed} & 0.8089   & 0.7270   & 1.8939 & 0.9194 & 0.9470 \\
\bottomrule
\end{tabular}
% }
\end{table}

First, we examine the performance of temporal-encoder-only and affinity-encoder-only models.
Both architectures are capable of identifying the merchant category of a merchant with various successes, with the temporal-encoder-only outperforming affinity-encoder-only.
When examining the performance of the fusion model with the simple-concatenate architecture shown in Figure~\ref{fig:simple}, although the improvement is marginal, the simple-concatenate architecture outperforms the temporal-encoder-only architecture.
This observation shows that even with relatively simple architecture, the idea of combining the merchant affinity information with merchant time series information does help us build a more robust merchant category identification model.
Lastly, we compare the proposed architecture (i.e., Figure~\ref{fig:overall}) with simple-concatenate architecture.
The proposed architecture outperforms the simple-concatenate architecture significantly (two sample t-test with a significance level of 1\% on Micro F1, Macro F1, and Hit@3).
The superb performance of the proposed method presented in this section combining with the visualization presented in Section~\ref{sec:affvis} confirms our assumption about the data: a better decision boundary can be learned between merchants' temporal representations when only considering merchants targeting similar set of costumers.
Generating logistic regression based on the affinity representations improves our system's capability of predicting the correct merchant category.

\subsection{Detecting Bad Actors}
\begin{figure}[ht]
    \centering
    \includegraphics[width=0.85\linewidth]{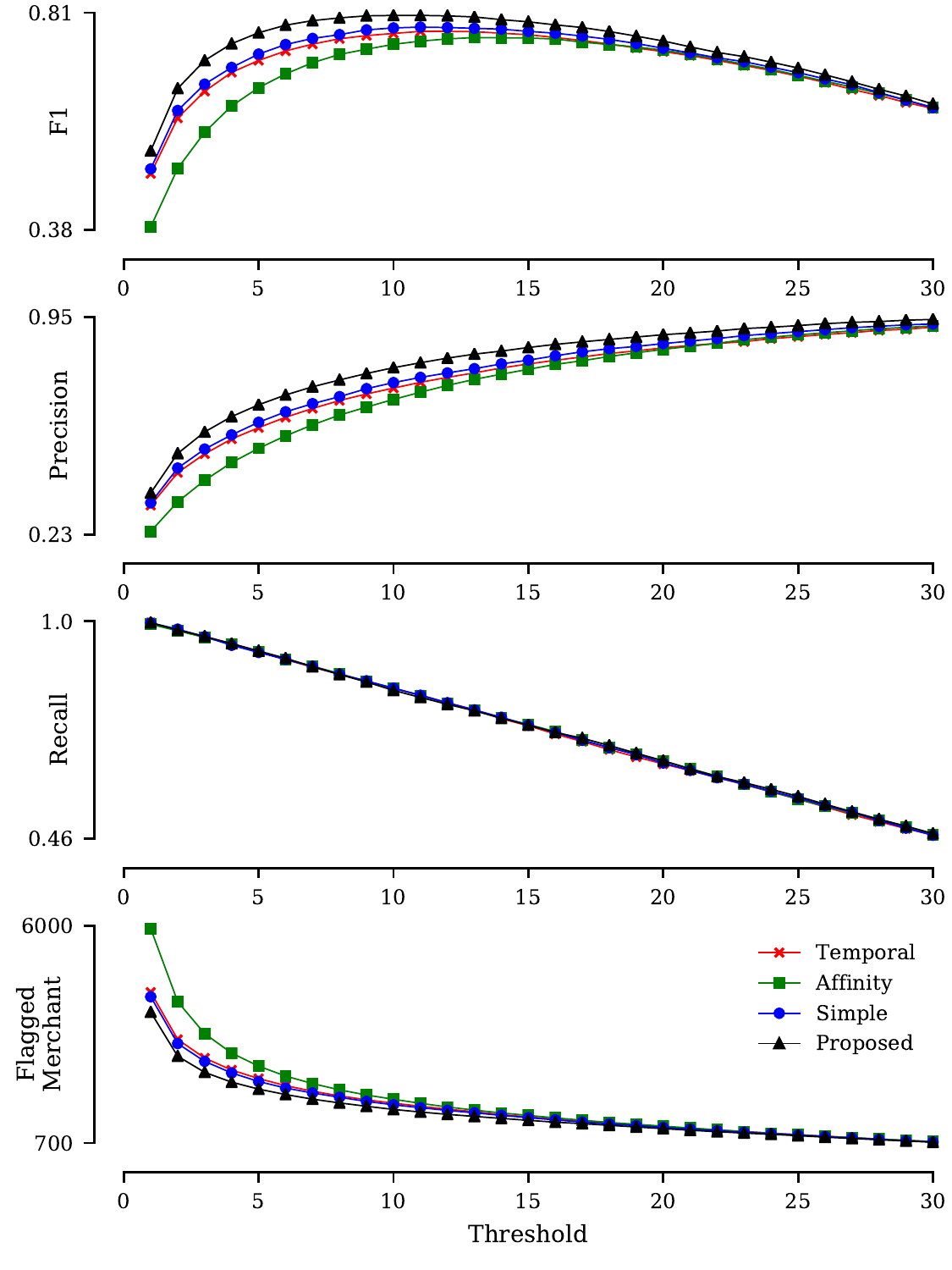}
    \caption{Performance on detecting merchants with false merchant category.}
    \label{fig:real_exp}
\end{figure}

As the actual business problem we want to solve is identifying merchants who report false merchant category to payment company (i.e., identifying \textit{bad actors}), we modify our dataset to test out the proposed method's capability of detecting such merchants.
We randomly select $10\%$ of the merchants in test data and randomly change their category label.
We label the selected merchants as bad actors, and the goal is to detect these bad actors.
To use the proposed system to identify the bad actors, we need to design a detection rule.
Here, the rule we use is: if the merchant category reported by a merchant is not in the top $k$ most probable categories, a bad actor is detected.
The $k$ is a threshold variable used to trade-off precision with recall.
Figure~\ref{fig:real_exp} shows the F1 score, precision, recall, and the number of merchants flagged as bad actors as we vary the threshold of $k$.
We use 5-fold cross-validation as in previous experiments.

First of all, similar to the conclusion of Table~\ref{tab:fusion}, the proposed method outperforms all the other alternatives.
When the threshold is smaller, merchants are more likely to be flagged as a bad actor because the criteria for a merchant being a bad actor is stricter.
Therefore, the precision is low due to many false positives, and the recall is high.
Alternatively, the precision would be high while the recall suffers when the threshold is high.
If we consider both precision and recall (i.e., F1 score), the optimal threshold would be $10$, and the corresponding precision, recall, number of the flagged merchants is $0.80$, $0.81$, $1455$ respectively using the proposed architecture.
In other words, $81\%$ of the bad actors are caught with the cost of investigating $291$ falsely flagged merchants.
For the best alternative (i.e., Simple) to achieve the same recall, $100$ more innocent merchants are falsely flagged as bad actors.
Note that the $k$ is set based on the bandwidth of the investigation team (i.e., number of flagged merchants) for deployment.

Aside from comparing the conventional performance measurements with the experimental dataset, it is also important to discuss the performance improvement in terms of business value in a more realistic setting.
Note that we can only use rough numbers in this discussion due to the confidential nature of the information, such as the actual number of merchants on a payment company's network.
Here, we assume a million merchants are using a payment network based on the estimated number of retail merchants in the U.S. from the National Retail Federation (NRF)~\cite{nrf}.
Because NRF only counts retail stores in the U.S. (i.e., online and non-U.S. merchants are ignored), this is a pessimistic estimation.
Assuming the investigation department of the payment company uses the same $k=10$ threshold, given that the size of the test set is around $14,334$ merchants in the above experiment if we scale the result reported in Figure~\ref{fig:real_exp} up to the number provided by NRF, the investigation department would investigate $6,900$ less false positive merchants using the proposed architecture comparing to the simple-concatenate architecture.
If each case needs an hour to investigate, the proposed architecture will save the investigators' working time by $287$ days comparing to the best alternative architecture.
As a result, the proposed architecture can significantly reduce the operation cost associated with the faulty merchant category investigation.

\section{Related Work} % File 6/7
There are hundreds of time series classification algorithms proposed over the years~\cite{bagnall2017great}.
Among them, the nearest neighbor classifier coupled with different time series distance measurements~\cite{UCRArchive2018}~\cite{lines2015time} remains to be a strong yet straightforward baseline, and the ensemble method (i.e., HIVE-COLT~\cite{bagnall2017great}) is considered the state-of-the-art time series classification method.

Due to the success of deep learning in domains of computer vision and natural language processing~\cite{krizhevsky2012imagenet,mikolov2013distributed,he2016deep,devlin2018bert}, there are many attempts in adopting deep learning-based models into time series classification or regression~\cite{cui2016multi,le2016data,tanisaro2016time,zheng2016exploiting,wang2017time,zhao2017convolutional,serra2018towards,yeh2020multi,zhuang2020multi}.
The majority of the methods mentioned above are variants of deep convolutional network~\cite{fawaz2019deep}.
For example, Cui et al.~\cite{cui2016multi} used smoothing and sampling techniques to capture discriminate information from time series in multiple scales and frequencies.
Le Guennec et al.~\cite{le2016data} modified a classical image classification convolutional network (i.e., LeNet~\cite{lecun1998gradient}) for time series classification.
Zheng et al.~\cite{zheng2016exploiting} proposed a two-stage convolutional network where the first stage focuses on extracting per channel (or dimension) features while the second stage learns to combine features extracted from each channel.
Wang et al.~\cite{wang2017time} demonstrated the effectiveness of a convolutional network with residual connection~\cite{he2016deep} in time series classification.
Zhao et al.~\cite{zhao2017convolutional} proposed another convolutional network-based model with unconventional pooling and non-linear components, and Serr{\`a} et al.~\cite{serra2018towards} examined the possibility of combining convolutional network-based model with attention mechanism~\cite{vaswani2017attention} in time series classification.

Based on recent benchmarks on time series classification and general sequence modeling~\cite{bai2018empirical}~\cite{fawaz2019deep}, convolutional network with residual connection~\cite{he2016deep} excels in both tasks.
As a result, we also adopt a variant of convolutional network with the residual connection in our temporal encoder design.
By comparing our temporal encoder design with other designs adopting other commonly used sequence model~\cite{bai2018empirical}, we have established the state-of-the-art method for modeling time series extracted from credit card transaction records.

Aside from modeling the temporal information generated by merchants, we also model the interaction between the merchants and consumers similar to~\cite{yeh2020towards}.
Interaction between different entities can generally be modeled as a graph~\cite{cai2018comprehensive, goyal2018graph,wu2020comprehensive}.
Many methods are proposed to model a graph by learning embedding representation for each node.
DeepWalk~\cite{perozzi2014deepwalk} learns embeddings by combining random walk and word2vec algorithm~\cite{mikolov2013distributed}.
Line~\cite{tang2015line} uses an edge-sampling algorithm to obtain the embeddings efficiently, and node2vec~\cite{grover2016node2vec} improved upon DeepWalk by using the biased random walk based on the principles in network science.
Graph convolutional network~\cite{kipf2016semi} applies the convolution operation on a graph to capture structure information from the graph.
As methods mentioned above do not work in the case of inductive learning (i.e., modeling nodes that were not seen during training time), Hamilton et al.~\cite{hamilton2017inductive} proposed the GraphSage algorithm which uses convolution operation to model graphs in an inductive learning scenario.
Because the merchant category identification problem requires inductive learning, we adopted the main idea presented in~\cite{hamilton2017inductive} to model the affinity between merchants.

To the best of our knowledge, our work is the first to present a model architecture that utilizes both time series and graph information for identifying a merchant's true category with credit card transaction history.

\section{Conclusion} % File 7/7
To accurately verify a merchant's self-reported merchant category, the proposed system utilizes two types of important information from transaction data: 1) the interaction between merchants and consumers and 2) the dynamics of the merchants' transaction activities.
We use an affinity encoder to capture the interaction information and a temporal encoder to capture the dynamics of a merchant.
To combine both types of information, we use a novel architecture design based on the assumption that consumers behave differently in different market segments.
The experiment result has shown the proposed method outperforms the alternative designs and provides great business values to the payment industry.
We are currently integrating this model into Visa's deep learning-based fraud detection system.

\bibliographystyle{IEEEtran}
% \bibliography{sections/ref.bib}

\end{document}